\documentclass[runningheads]{llncs}
\usepackage[T1]{fontenc}
\usepackage{graphicx}
\usepackage{multicol}
\usepackage{footmisc}
\usepackage[natbib,style=numeric, minbibnames=5, maxbibnames=99, sorting=none, sortcites=true, maxsortnames=99, minsortnames=1]{biblatex} %
\usepackage{graphicx} 
\usepackage{amsmath,amssymb,amsfonts}
\usepackage{hyperref}
\usepackage{booktabs} 
\usepackage{multirow}
\usepackage{xcolor}
\usepackage{mleftright}
\mleftright 

\bibliography{ref.bib}
\title{LIAM: Multimodal Transformer for Language Instructions, Images, Actions and Semantic Maps}
\titlerunning{LIAM: Multimodal Transformer for Language, Images, Actions and Maps}
\author{Yihao Wang, Raphael Memmesheimer, Sven Behnke}
\institute{Autonomous Intelligent Systems, Computer Science Institute VI, Center for Robotics, University of Bonn, Germany\\
\url{https://ais.uni-bonn.de}
}
\begin{document}
\maketitle
\begin{abstract}
The availability of large language models and open-vocabulary object perception methods enables more flexibility for domestic service robots. The large variability of domestic tasks can be addressed without implementing each task individually by providing the robot with a task description along with appropriate environment information.
In this work, we propose LIAM --- an end-to-end model that predicts action transcripts based on language, image, action, and map inputs. Language and image inputs are encoded with a CLIP backbone, for which we designed two pre-training tasks to fine-tune its weights and pre-align the latent spaces. We evaluate our method on the ALFRED dataset, a simulator-generated benchmark for domestic tasks. Our results demonstrate the importance of pre-aligning embedding spaces from different modalities and the efficacy of incorporating semantic maps.
\end{abstract}
\section{Introduction}
With the rapid evolution of deep learning research, particularly in the natural language domain, we have witnessed the emergence of enormous Transformer-based models capable of generating high-quality texts and synthetic images~\cite{brown2020language, ramesh2022hierarchical}. They hold much potential for open-ended robotic applications, like domestic service tasks, but cannot be directly applied because robotics involves additional modalities, such as images and actions.

In 2021, OpenAI released the foundation model CLIP, which uses a contrastive learning paradigm to connect text and images~\cite{radford2021learning}. Representations of textual and visual inputs are learned by contrasting similar and dissimilar pairs. 
CLIP features are suitable for open-vocabulary zero-shot image categorization.

Many robotics applications soon utilized CLIP-based models for different open-vocabulary tasks~\cite{shafiullahclip}. 
Vision-Language-Action models like OpenVLA~\cite{OpenVLA} predict the next action based on visual inputs.
Research in models predicting entire action transcript sequences based on language instructions and visual inputs is still in its early phase. Training a model that understands all relevant modalities and has strong generalizability is challenging and requires a considerable amount of data. 

In this work, we introduce the end-to-end model LIAM, which receives inputs from the modalities \textbf{L}anguage, \textbf{I}mages, \textbf{A}ctions, and \textbf{M}aps and predicts action sequences. Two CLIP-like self-supervised training paradigms are designed to pre-align two or more different embedding spaces during the pre-training stage, enhancing the model's ability to understand and process multimodal inputs. We utilized semantic maps as an additional modality directly in addition to textual and visual input, where the map modality is designed to serve as a knowledge base for the model. We evaluate our model with the ALFRED Challenge~\cite{shridhar2020alfred}, a benchmark task that requires a robot to understand and execute complex natural language instructions in a simulated environment. 
The code and pre-trained models to reproduce the results are made publicly available: \url{https://github.com/AIS-Bonn/LIAM}.

\section{Related Work}
Open-vocabulary approaches have garnered significant attention in the field of computer vision. In the past decade, most vision downstream tasks were trained via datasets with a fixed number of classes. The CLIP model, with its contrastive learning paradigm, brings a promising solution for inferencing open vocabulary vision tasks. After embedding the text and image using separate encoders, CLIP is trained by maximizing the similarity score of the representation of corresponding image and text pairs~\cite{radford2021learning}. 

A massive amount of work is built on CLIP models for different down-streaming tasks, e.g., object detection~\cite{zhou2022detecting, gu2021open}, semantic segmentation~\cite{luddecke2022image, xu2022groupvit}, action prediction~\cite{wang2021actionclip, luo2022clip4clip}. These approaches have designed similar methods adapted to their specific vision downstream tasks. For example, GroupViT first uses a group-based approach for semantic segmentation by introducing learnable group tokens to the ViT architecture (\cite{xu2022groupvit}). During inference time, each group is assigned the semantic label with the highest similarity score representing the text input. \\
To tackle the Vision-Language Navigation (VLN) and Vision-Language Manipulation (VLM) tasks, there are two main approaches. I. An End-to-End model trains a model that can accept all modal inputs simultaneously and make predictions of current action for the agent~\cite{shridhar2020alfred, pashevich2021episodic}. For example, Episodic Transformer (E.T.) is an end-to-end attention-based transformer model that consists of three modality-specific encoders and one multi-modality fusion encoder \cite{pashevich2021episodic}. II. Modular Methods propose a pipeline of learned modules for tackling the challenge of understanding the environment and the tasks~\cite{blukis2022persistent, min2021film}. 
FILM proposes a language processing module, which converts the task description into an action transcript using pre-defined templates. A semantic mapping module processes egocentric RGB images into a 2D semantic map. Furthermore, a semantic search policy predicts the possible distribution of objects’ locations. A deterministic policy is designed to predict the action decision based on the predicted action transcript and the predicted semantic map. 

Most end-to-end models for VLM tasks primarily use language and images as inputs. In this paper, we also generate a semantic map following the FILM approach and incorporate it as an additional modality input.

\section{Main Approach}
\begin{figure}[t]
    \centering
    \includegraphics[width=\linewidth]{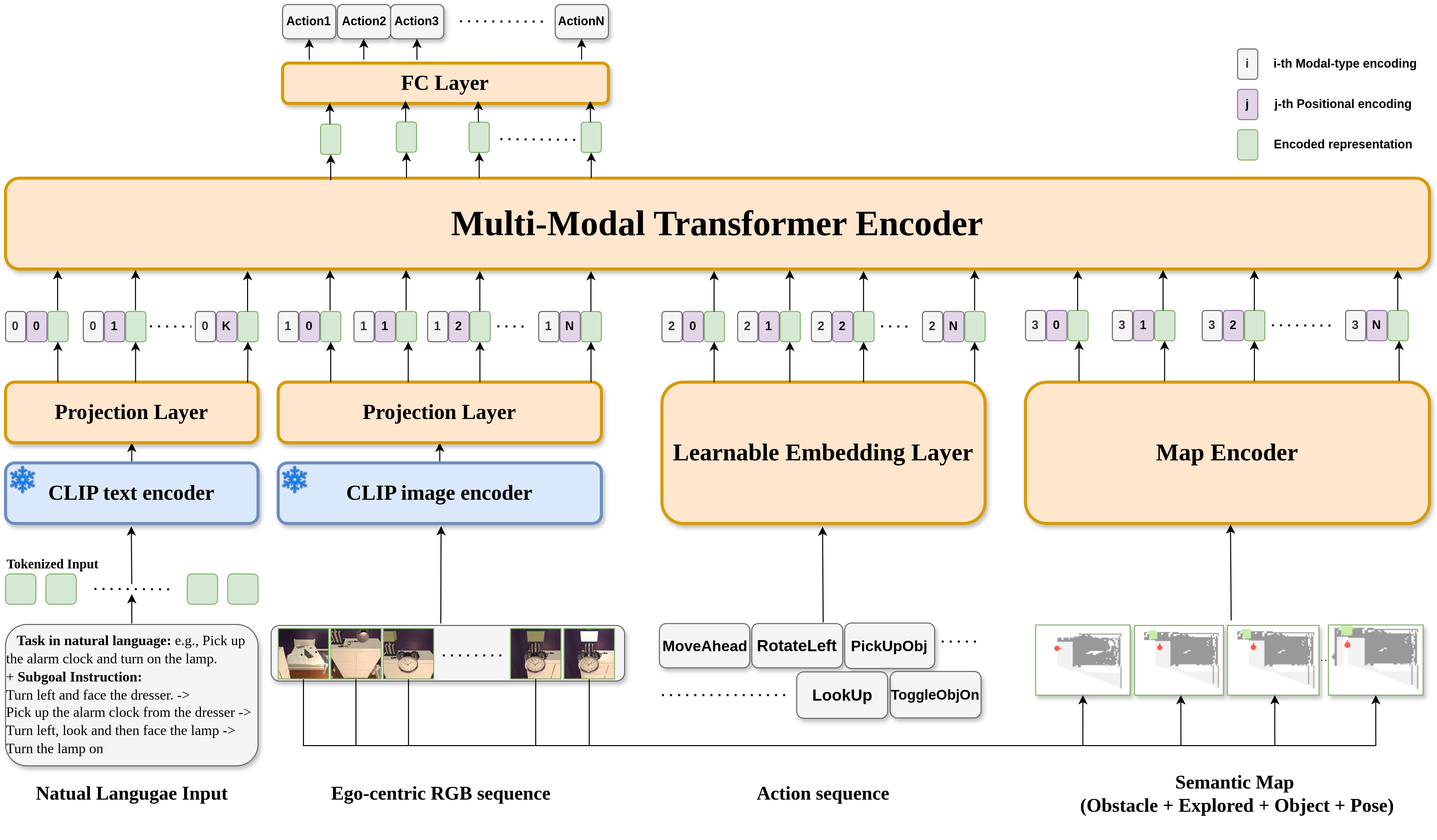}
    \caption{Model architecture of LIAM. The blue blocks are all layers that were frozen during the end-to-end model training; the orange blocks are the parts that were trained. }
    \label{fig:approach overview}
\end{figure}
We now introduce the LIAM model, including the pre-training stage and the action generation. An overview of the model is given in Figure \ref{fig:approach overview}. The model architecture is adopted from the Episodic Transformer~\cite{pashevich2021episodic}, but we replace the image and text encoders with CLIP encoders and employ a semantic map~\cite{min2021film}. To manage the high computational cost, we freeze the weights and biases of the CLIP backbone during end-to-end model training, ensuring efficient resource utilization. \\

All experiments are conducted using ALFRED (Action Learning From Realistic Environments and Directives)~\cite{shridhar2020alfred}, a recent simulator generated benchmark for learning a mapping task, from vision (egocentric RGB image) and natural language input to an action transcript for a domestic service robot.
ALFRED's challenge provides the following annotations: an initial state of the simulated room, language instructions, and an expert demonstration trajectory. 
Language instructions are provided as global instructions, e.g., "Pick up the alarm clock and turn on the lamp. $<<$goal$>>$" and a list of sentences for sub-goal instructions.
The annotation consists of a sequence of discrete actions, the object's mask, whenever the interaction with objects is involved. \\

\begin{figure}[ht]
    \centering
    \begin{minipage}[t]{0.45\textwidth}
        \centering
         \includegraphics[width=\textwidth]{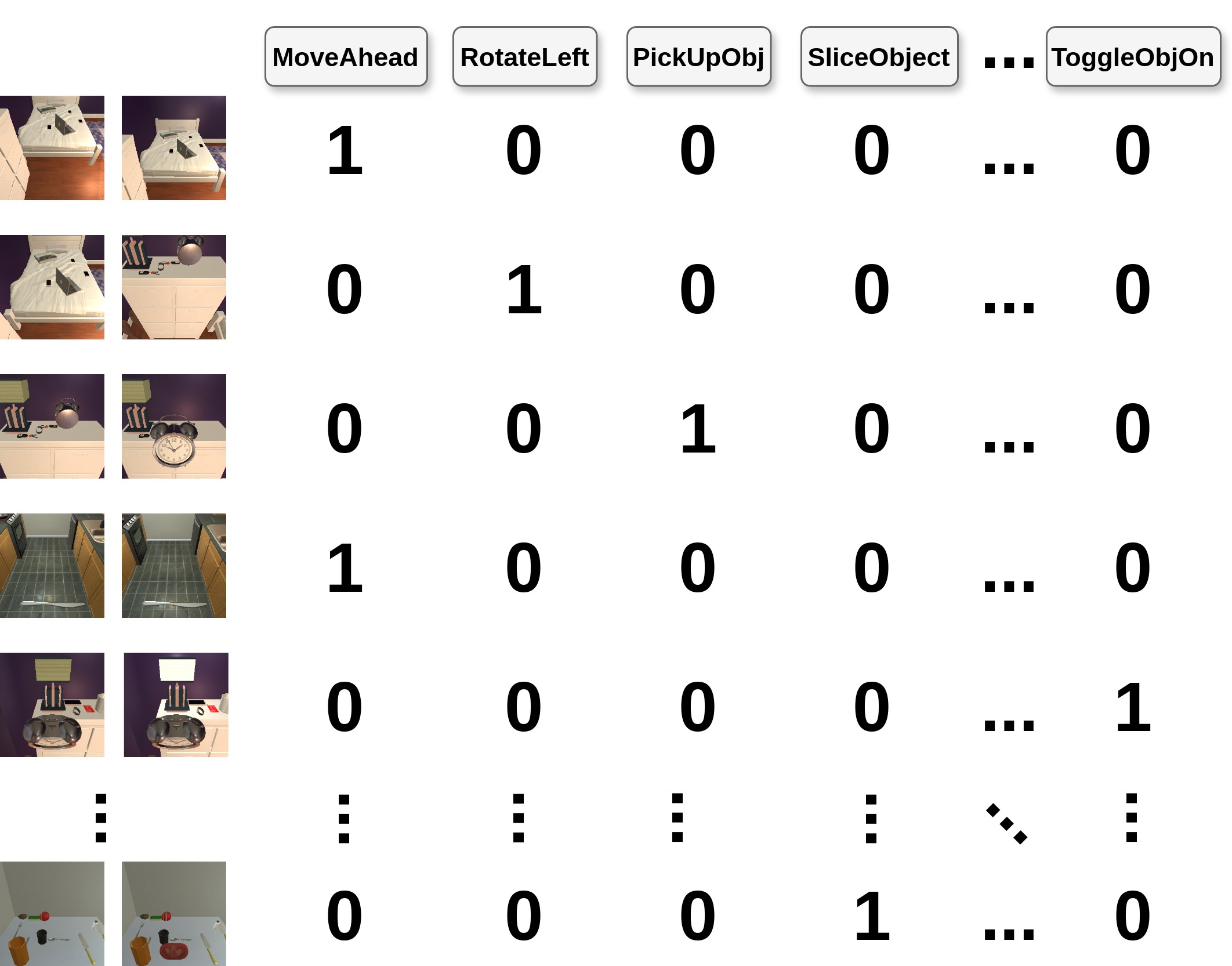}
         \caption{An example of the ground truth from one mini-batch for alignment of visual and action embedding space. One action corresponds to two consecutive frames.}
    \label{fig:pre_gt1}
    \end{minipage}
    \hfill
    \begin{minipage}[t]{0.5\textwidth}
        \centering
        \includegraphics[width=\textwidth]{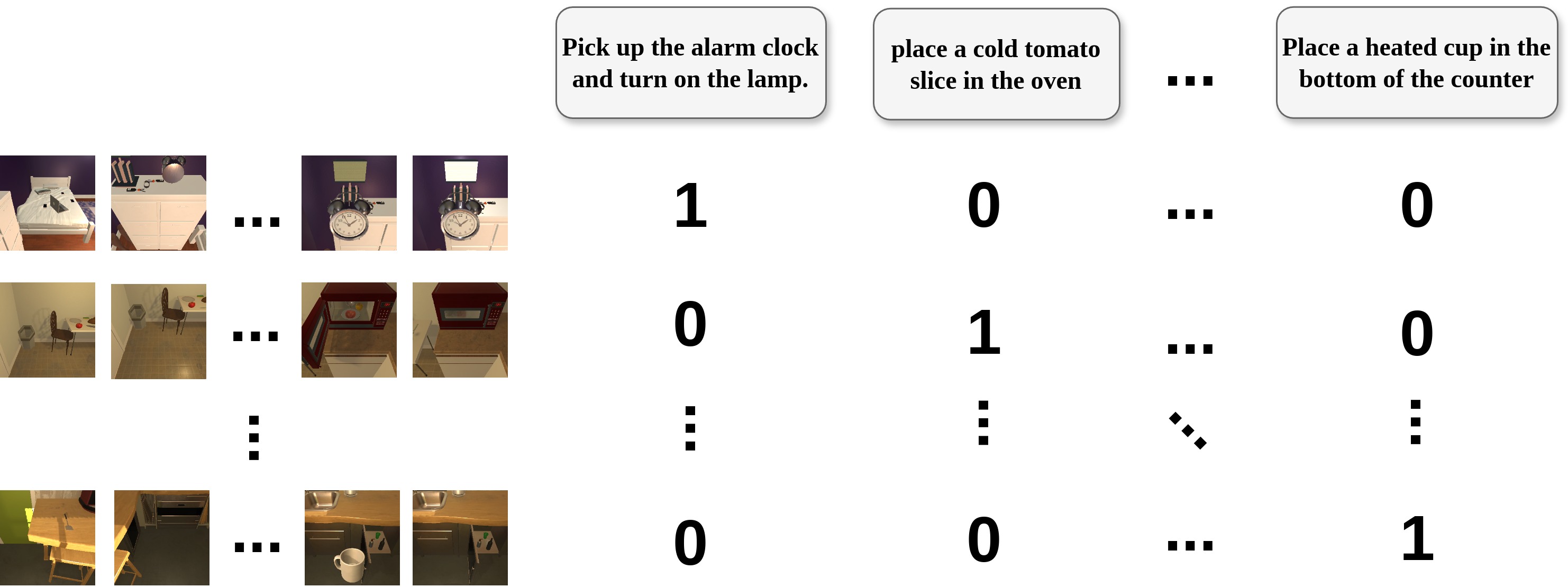}
        \caption{Ground truth for alignment of visual and language embedding space. One frame sequence corresponds to one language instruction.}
        \label{fig:pre_gt2}
    \end{minipage}
    \vspace{-0.4cm}
\end{figure}

\subsection{Contrastive Alignment Pre-training} \label{ch:basiccon}
CLIP was initially trained using data gathered from the Internet, resulting in bias and significant performance differences across different datasets in zero-shot inference~\cite{radford2021learning}. Given that the ALFRED dataset is simulator-generated, it deviates slightly from reality. Hence, a pre-training stage to align the latent space of image and action becomes imperative. \\
We first pre-aligned the CLIP image encoder and action embedding via contrastive learning. The learning objective of the model is to match the correct pair of action embedding and image embedding. Unlike CLIP, where one image has one corresponding label, in our case, two consecutive images have one corresponding label, which denotes an action. Figure \ref{fig:pre_gt1} shows an example of the ground truth of one mini-batch. Unlike the original CLIP, which forms an N $\times$ N affinity matrix (N denotes the batch size), our approach counts the number of unique actions that appear in each mini-batch instead and forms all the actions occurring as the column of the matrix. Each row denotes the representation of two consecutive frames. The matrix entry (i, j) has the value one if the i-th frame's representation corresponds to the action class j.

In our approach, we used f$(\cdot, \cdot)$, a 1D convolution layer along with global average pooling, to compute a fused representation of two consecutive images. We embedded actions with an embedding layer that embeds all 14 classes (12 classes + 2 special tokens) into 768-dimensional dense vectors. After having the normalized representations of both frames $I_e$ and action classes $A_e$, we computed a cosine similarity score $s$ of both vectors and fine-tuned the CLIP backbone following CLIP's approach. With $a_i \in A_e, I_t \in I_e$, the similarity matrix P is defined as:

\begin{equation*}
       P_\textrm{I2A}\left(I\right) = \frac{\exp\left(s\left(f\left(I_t, I_{t+1}\right), a_i\right) / \tau\right)}
       {\sum\limits_{j=1}^{N - 1}e^{s\left(f\left(I_t, I_{t + 1}\right), a_j\right) / \tau}}, 
 \quad P_\textrm{A2I}\left(A\right) = \frac{\exp\left(s\left(a, f\left(I_{t_i}, I_{t_{i}+1}\right)\right) / \tau\right)}{\sum\limits_{j=1}^{N - 1}e^{s\left(a_j, f\left(I_{t_i}, I_{t_{i} + 1}\right)\right) / \tau}}.
\end{equation*}



We notated text, action, and image with the letters T, A, and I. I2A denotes "image-to-action," and all later mentioned abbreviations (X2X) follow this pattern. Function f$(\cdot, \cdot)$ denotes the fusion function of two consecutive images at time step t and t + 1 as described above. $\tau$ is the temperature, a learnable parameter scales the similarity score. Following CLIP, we initialized the temperature with a value of 0.07 and use 100 as an upper threshold to clip the temperature value during the training stage, preventing the scaling factor from becoming too large~\cite{wang2021actionclip} and 0.01 as a lower threshold bound to prevent dividing by zero error. 

Following ActionCLIP~\cite{wang2021actionclip}, we used the KL divergence loss to fine-tune the CLIP backbone. The total loss of the contrastive alignment pre-training is the average of both image-to-action and action-to-image loss: 
$$\mathcal{L}_{\text{Image-Action}} = \frac{1}{2}\mathbb{E}_{(x,y)\backsim \mathcal{D}}\left[KL\left(P_{\text{I2A}}\left(I\right), Q_{\text{I2A}}\left(I\right)\right) + KL\left(P_{\text{A2I}}\left(A\right), Q_{\text{A2I}}\left(A\right)\right)\right],$$ where $Q(\cdot)$ denotes the ground truth similarity matrix. 

\subsection{Triple Contrastive Pre-training}
Our contrastive learning pre-aligns the latent space of the vision and action embedding. However, the alignment of the language embedding space with other latent spaces is neglected. To address this, we introduced a triple contrastive pre-training stage. This method not only aligns the vision and action embedding spaces but also aligns the language embedding space with them. 
This mutual training ensures that the loss can be back-propagated to all encoders. 
In addition to Figure \ref{fig:pre_gt1}, Figure \ref{fig:pre_gt2} shows the example of the ground truth of aligning the CLIP text encoder and image encoder. 
The training objective is straightforward: the representation of the language instruction and its corresponding frame sequence should be similar to each other. We first computed the representation of every single frame (N frames) using the CLIP image encoder and further processed this sequence of representations in two ways. 

Considering the computing resource for the image sequence representation, we chose the parameter-free approach to compute the mean value of all CLIP embeddings of all frames to get the sequence representation. For every two consecutive frames in this list, we used the 1D convolution layer to get the representation of both frames as the same methodology described above in Section \ref{ch:basiccon}. Thus, the visual embedding space is aligned simultaneously with both text and action embedding space. As shown below, we define cross-entropy loss $\mathcal{L}_{\text{Text-Image}}$ for learning the bidirectional pairing of visual and text sequences following the original CLIP. For aligning action and image, we continued using KL divergence loss $\mathcal{L}_{\text{Image-Action}}$discussed in Section \ref{ch:basiccon}: 
\begin{equation*}
\scalebox{0.99}{$\mathcal{L}_{\text{Text-Image}} = -\frac{1}{2N} \Bigg(  \sum^N_{i=1} \log\left(\frac{\exp(S_{ii} / \tau)}{\sum_{j=1}^N \exp(S_{ij} / \tau)}\right)
+  \sum^N_{j=1} \log\left(\frac{\exp(S_{jj} / \tau)}{\sum_{j=1}^N \exp(S_{ij} / \tau)}\right) \Bigg),$}
\end{equation*}

where S is the similarity matrix.
The total loss of triple contrastive alignment is defined as:
$$\mathcal{L}_{\text{total}} = (1 - \alpha) \times \mathcal{L}_{\text{Text-Image}} + \alpha \times \mathcal{L}_{\text{Image-Action}}.$$
We introduced $\alpha$ as a hyperparameter for weighing both losses. This hyperparameter gives us more flexibility. We assigned ($\alpha = 0.8$) in the total loss. 
\subsection{Action Generation}
We embedded text and visual input using independent CLIP encoders, which are now pre-aligned. The map is embedded using a trainable projection layer (FC layer + GELU). We first applied positional encoding and modal-type encoding to each representation. The learnable modal-type encoding, first introduced by VILT \cite{kim2021vilt}, brings extra information and clarifies different modality types to the Transformer model; it also avoids the need for explicit separation tokens like “[SEP].” \\ After all the representations are concatenated, we fed it further to a multi-modality fusing layer (2-layer Transformer encoder) to learn a general representation. 
The multimodal Transformer is a comprehensive model that aims to integrate all different modalities into one global representation, namely language, visual frames, action sequence, and semantic map in our case. We use causal attention following E.T. \cite{pashevich2021episodic}. The attention mask follows these rules: Language tokens should attend to the language itself. Visual frames are allowed to attend to all language tokens, but they can only attend to all the frames, actions, and semantic maps before the current time step t. The same rules are applied to action tokens and semantic maps. Sinusoidal positional encoding is also applied.  

Ultimately, we employed a fully connected layer to predict an output sequence based on all the tokens of visual representations, which have the same number of mappings of actions so that we can train the model with the ground truth action sequence. 

\subsection{LIAM Pipeline}

In this section, we define our input and output mathematically to explain our approach more deeply. 
Given the following inputs: Tokenized language instructions $[L_1, … L_m]$, 
image sequence $[I_1,...,I_n]$, action sequence $[A_1, …, A_n]$, semantic map sequence $[M_1, …, M_n]$, where 
language tokens have length m, and all other modalities have length n. For an image sequence $I_1, \dots, I_n$, only n - 1 actions are involved. We append "<<stop>>" as the last action. Four different encoders encode all input to their latent space and output $F_l, F_v$,$F_a$ and $F_m$ as the feature representation: 
\begin{align*}
&\overline{F}_{mod} = F_{mod} + mod^{Type} + mod^{Pos}, \hspace{2mm} \text{for mod in L, I, A, M.} \\
&F_{total}^0 = [\overline{F_L};\overline{F_I};\overline{F_A};\overline{F_M}] \\
&\overline{F}^l_{total} = LN(MA(\overline{F}_{total}^{l-1})) + \overline{F}^{l - 1}_{total},  \hspace{22mm} l=1,...,\mathbb{L} \\
&\overline{F}^l_{I} = \overline{F}^d_{total}[m + 1: m + 1 + n] \\
&O = \overline{F}^d_{I}D.
\end{align*}
After applying modal-type and positional encoding to each feature representation, we concatenate them into $F_{total}^0$. The multi-modal Transformer encoder has $\mathbb{L}$ layers, each with Multi-headed Attention (MA), Layer Normalization (LN), and residual connections. We then extract the visual part of the total representation (from index m + 1 to m + 1 + n) for action sequence prediction. The final action sequence is obtained by multiplying this sliced matrix with a trained linear layer D.
The end-to-end model is trained using cross-entropy loss between the generated action sequence and the ground truth action sequence. For predicted action sequence P = [$\hat{a_1} \dots \hat{a_n}$] and ground truth Q = [$a_1 \dots a_n$], the cross-entropy loss is defined as: 
$$\mathcal{L}_{\text{action}} = - \sum^n_{i=1}\sum^{13}_{c=1} a_{i,c}log(\hat{a_{i,c}}), $$ where c represents the action class out of 14 classes (12 classes + <<stop>> + <<pad>>). However, the <<pad>> token is masked out during the computation of the loss value. \\
Following E.T. \cite{pashevich2021episodic}, we also use auxiliary losses in the training. First, we predict the object class for each frame ("NoObject" is also a class). The predicted object class is a list with the same length as the action. Similary to the action, for predicted object sequence $P_o$ = [$\hat{o_1} \dots \hat{o_n}$] and ground truth Q = [$o_1 \dots o_n$], the cross-entropy loss is defined as: 
$$\mathcal{L}_{\text{object}} = - \sum^n_{i=1}\sum^{85}_{c=1} o_{i,c}log(\hat{o_{i,c}}), $$ where c represents the object class out of 85 classes (84 object classes + NoObject).\\
We also predict the goal progress with the initial thoughts for giving the model a better understanding of the current progress of the current trial. The ground truth $Q^{gp}$ of the goal progress is computed by $$ Q^{gp} = \left\{ \frac{i + 1}{n} \mid i \in \{0, 1, 2, \ldots, n-1\} \right\} ,$$ where n is the length of the ground truth action for the current trial. For predicted goal progress $P^{gp}$, we use Mean Square Error (MSE) loss, defined as:
$$\mathcal{L}_{gp} = \frac{1}{n} \sum_{i=1}^n (Q^{gp}_i - P^{gp}_i)^2.$$
The total loss is then defined as: 
$$\mathcal{L}_{total} = \mathcal{L}_{action} + \alpha \mathcal{L}_{object} + \beta \mathcal{L}_{gp},$$ where $\alpha, \beta$ are hyperparameters for weighing the auxiliary loss. We set all weights to 0.1 in our experiments.

\section{Experiments}

We now present results from the pre-training stage, and report quantitative and qualitative results.

\begin{table}[t!]
\caption{Matching accuracy of image-to-text (I2A) and text-to-image (T2I) pairs.}
\centering
\resizebox{\textwidth}{!}{%
\begin{tabular}{lccccc}
\toprule
\multirow{2}{*}{\textbf{Models}} & \multicolumn{2}{c}{\textbf{Validation}}&  & \multicolumn{2}{c}{\textbf{Test}}\\
        \cmidrule{2-3} \cmidrule{5-6}      
        {} & ACC (I2A) & ACC (T2I) & & ACC (I2A)  &  ACC (T2I)  \\
\midrule
\textbf{Contrastive alignment of image and action}  & & & & \\  
\midrule
  MobileCLIP S0 (Zero-shot) &  7.51 & -  & & 7.32     & -    \\  
  MobileCLIP S0 & 38.34  & - &  & 43.21  & -    \\  
  OpenCLIP-RN50 (Zero-shot) & 5.63 & - & & 5.51 & - \\ 
  OpenCLIP-RN50 &  61.41   & - &     & 58.78& - \\  
  OpenCLIP-ViT-B-32 (Zero-shot) & 5.84 & - & & 8.26   & -    \\  
  OpenCLIP-ViT-B-32 & \textbf{65.45} & -    && \textbf{61.58} & -    \\  

\midrule
\textbf{Triple contrastive alignment} & & & & & \\  
\midrule
  MobileCLIP S0 (Zero-shot) & 3.21 & 35.04 & & 3.34  & 33.03 \\  
  MobileCLIP S0 & 80.28 & 60.22 & & \textbf{68.97} & 35.95 \\  
  OpenCLIP-RN50 (Zero-shot) & 3.76  & 31.51 &     & 4.78    & 34.61    \\  
  OpenCLIP-RN50 &   70.56  &  32.67   &    & 65.54 & \textbf{51.82}    \\  
  OpenCLIP-ViT-B-32 (Zero-shot) & 3.89 & 45.62 & &  5.67   & 37.65     \\  
  OpenCLIP-ViT-B-32 &  \textbf{88.22}   &  \textbf{75.43}  & & 65.98   & 40.69     \\  

\bottomrule
\end{tabular}}
\label{table:cont_align}
\end{table}

\subsection{Pre-training Stage}
We pick three different CLIP backbones: MobileCLIP-S0 (54.7 million parameters), OpenCLIP with Resnet-50 backbone (114.1 million parameters), and OpenCLIP with the vision transformer backbone (152.7 million parameters).

Table \ref{table:cont_align} presents the accuracy score of correct matching pairs among the mini-batches. Firstly, we can clearly see that no CLIP model can correctly match two consecutive visual images with their corresponding actions in a zero-shot manner. As discussed before, CLIP is a model that is sensitive to the dataset. However, both image-action contrastive and triple contrastive alignment can be adequately trained and show significant accuracy gains in matching corresponding pairs of two consecutive images and action or pairs of the image sequence and text sequence. We use batch size 64 for image-action contrastive alignment training, batch size three, and additionally, constraint each sequence to length 21 (during validation) for triple-contrastive alignment to keep a fair comparison between both accuracy scores. We observe that using zero-shot inference, the accuracy for text sequence and image sequence matching is around 33\%, and the accuracy for two image frames and action matching lies between 3\% and 8\%. Both scores denote the random guessing from the model. After fine-tuning, the accuracy of predicting the correct image and text sequence pair is increased using MobileCLIP backbone and CLIP VIT-B-32 backbone to 60.22\% and 45.62\%, however still not entirely correct out of the mini-batch. We observe that triple-contrastive alignment indeed benefits accuracy in matching the correct action label of the consecutive images using all three backbones. Overall, based on its performance in the action-image matching task, we decided to use pre-aligned CLIP VIT-B-32 as the backbone to extract the features of our input. 

\subsection{Quantitative Results}
\begin{table}[t!]
\caption{Results for the end-to-end model predicted sequence and the ground truth.}
\centering

\begin{tabular}{lccccc}
\toprule
\multirow{2}{*}{\textbf{Models}} &  \multicolumn{2}{c}{\textbf{Valid Seen}} & & \multicolumn{2}{c}{\textbf{Valid Unseen}}\\    
        \cmidrule{2-3} \cmidrule{5-6}      
{} & Accuracy  & F1 score & &Accuracy  & F1 score \\  
\midrule
Zero-shot CLIP encoder    & 73.00 &58.64  & &71.87&49.31    \\
+ Semantic Map &78.08 &71.69& &76.49&59.54           \\  
\midrule
Contrastive-aligned CLIP  &  76.97   &64.17 &&79.56&44.02           \\  
+ Semantic Map     & \textbf{79.30} &56.94 &&\textbf{81.51}&39.52        \\ 
\midrule
Triple-contrastive-aligned CLIP  & 77.52 & 72.15  & & 76.89&64.57      \\          
+ Semantic Map        &    77.98  & \textbf{72.63} &&75.79&\textbf{67.02}        \\ 
\bottomrule
\end{tabular}
\label{table:actionacc}
\end{table}

\begin{figure}[ht!]
    \centering
    \includegraphics[width=0.7\textwidth]{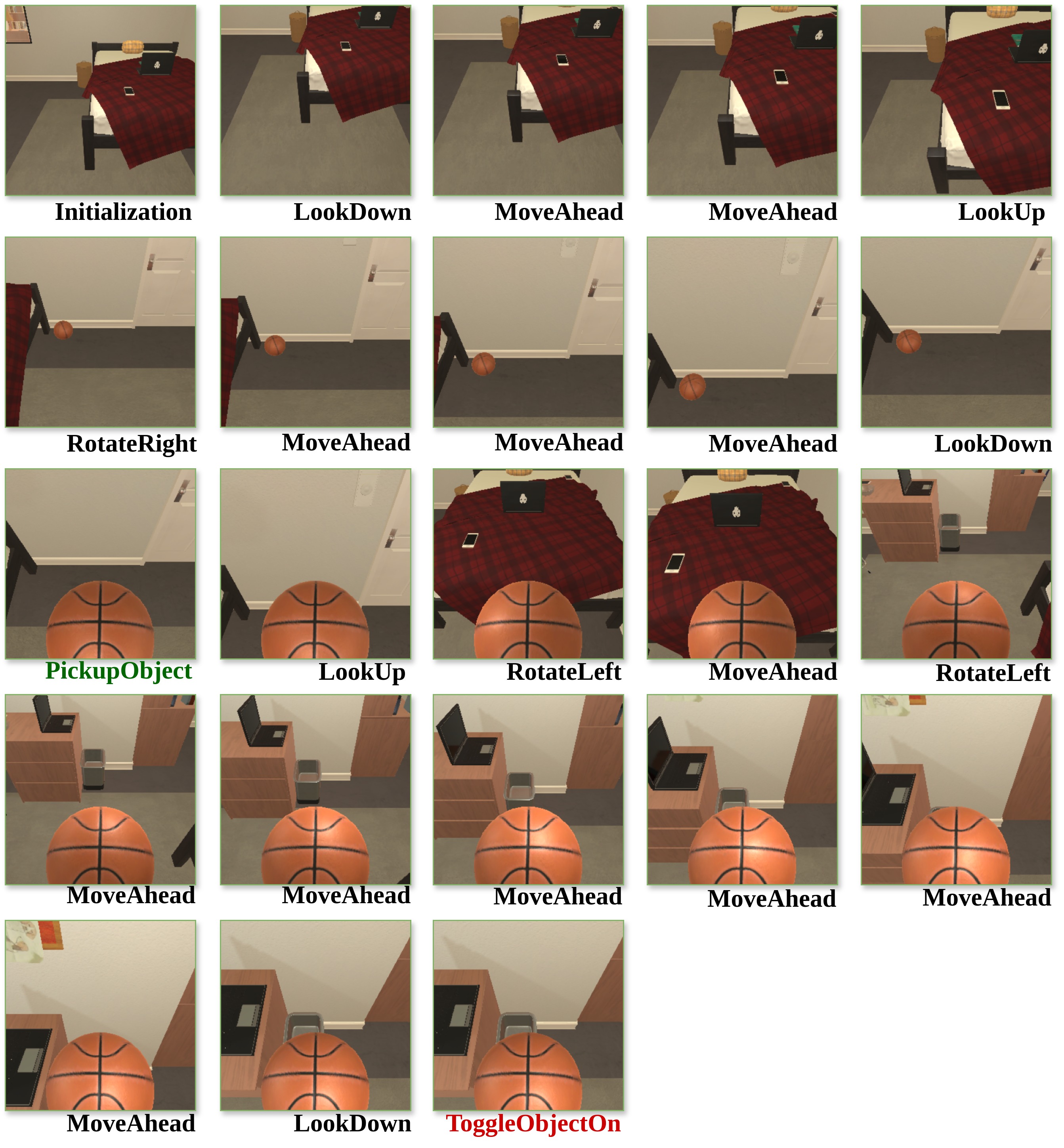}
    \vspace{-0.1cm}
    \caption{Example task: "Look at the basketball in the light from the lamp". The given step-by-step instructions are as follows: 'Walk to the foot of the bed.' -> 'Pick up the basketball from the floor.' -> 'Go to the desk to your left.' ->'Turn on the lamp.'}
    \label{fig:case1}
    \vspace{-0.3cm}
\end{figure}
In our experiments, we used OpenCLIP with ViT B-32 as our backbone, which we found to be the most suitable in our experiments. From Table \ref{table:actionacc}, first, we observe that the fine-tuning of the CLIP backbone brought benefits in increasing the accuracy and the F1 score. Although the improvement in accuracy is not significant, e.g., 73.00\%, 76.97\%, 77.52\% in the seen data, 71.87\%, 79.56\%, 76.89\% in the unseen data, respectively. When the triple contrastive pre-alignment is applied, the F1 score improves by 8\% on seen data and by around 20\% on unseen data, compared to the contrastive alignment approach. Second, the semantic map improved the model's performance for the zero-shot backbone and triple-aligned backbone, showing the potential of using the map as an extra modality to enhance the model's spatial understanding rather than solely given the egocentric RGB images. The backbone fine-tuned with image-to-action alignment has an acceptable accuracy score but a relatively low F1 score; this indicates that the model still suffers from the imbalanced data problem.

\subsection{Qualitative Results}
Figure \ref{fig:case1} shows the decision of our agent for the task, "Pick up the basketball and turn on the desk lamp in the bedroom." The agent can comprehend most of the sub-goal instructions and correctly find and pick up the basketball beneath the bed. Furthermore, the agent rotates left twice and goes to the desk. However, this task eventually failed because of the model's wrong prediction of the lamp's location (which should be on the other side of the desk).

\section{Conclusion}
In this paper, we introduced LIAM, a multimodal model incorporating natural language input, egocentric RGB image input, action history, and an accumulated semantic map. Because all the modality-specific encoders encode inputs in their own latent space, we designed two pre-training tasks to pre-align the embedding spaces. This pre-training objective aims to maximize the agreement of the action representation and the global representation of two consecutive images. However, the pre-training task does not consider the alignment of the text encoding backbone. We introduced a triple contrastive alignment to address this issue. The agreement between the image sequence representation and the text sequence representation, the two-frame representation and the action representation are expected to be maximized. 

Our model outperforms our baseline, using the OpenAI-released CLIP model and no map as an additional modality. We showed the importance of pre-aligning the embedding spaces from different modalities. In addition, using semantic maps as a modality to the end-to-end model brought benefits as well.

Future work holds immense potential for further enhancing our model. One avenue could be integrating the segmentation model inside our training stage and the semantic map generation stage. Furthermore, the methodology for encoding the map spatial information is another area that needs to be further investigated. We are particularly intrigued by the potential of considering maps as a knowledge base modality to the vision-language models.

\paragraph*{Acknowledgements:}
This work has been funded by the German Ministry of Education and Research under the grant no. 16SV8683, project: Transferzentrum Roboter im Alltag (RimA).
\printbibliography

@inproceedings{shridhar2020alfred,
  title={ALFRED: A benchmark for interpreting grounded instructions for everyday tasks},
  author={Shridhar, Mohit and Thomason, Jesse and Gordon, Daniel and Bisk, Yonatan and Han, Winson and Mottaghi, Roozbeh and Zettlemoyer, Luke and Fox, Dieter},
  booktitle={IEEE/CVF Conference on Computer Vision and Pattern Recognition (CVPR)},
  pages={10740--10749},
  year={2020}
}

@article{OpenVLA,
  author       = {Moo Jin Kim and
                  Karl Pertsch and
                  Siddharth Karamcheti and
                  Ted Xiao and
                  Ashwin Balakrishna and
                  Suraj Nair and
                  Rafael Rafailov and
                  Ethan Paul Foster and
                  Grace Lam and
                  Pannag Sanketi and
                  Quan Vuong and
                  Thomas Kollar and
                  Benjamin Burchfiel and
                  Russ Tedrake and
                  Dorsa Sadigh and
                  Sergey Levine and
                  Percy Liang and
                  Chelsea Finn},
  title        = {{OpenVLA}: An open-source vision-language-action model},
  journal      = {CoRR},
  volume       = {abs/2406.09246},
  year         = {2024},
}

@inproceedings{pashevich2021episodic,
  title={Episodic transformer for vision-and-language navigation},
  author={Pashevich, Alexander and Schmid, Cordelia and Sun, Chen},
  booktitle={IEEE/CVF International Conference on Computer Vision (CVPR)},
  pages={15942--15952},
  year={2021}
}

@inproceedings{blukis2022persistent,
  title={A persistent spatial semantic representation for high-level natural language instruction execution},
  author={Blukis, Valts and Paxton, Chris and Fox, Dieter and Garg, Animesh and Artzi, Yoav},
  booktitle={Conference on Robot Learning (CoRL)},
  pages={706--717},
  year={2022},
  organization={PMLR}
}

@inproceedings{min2021film,
  title={{FILM}: Following instructions in language with modular methods},
  author={Min, So Yeon and Chaplot, Devendra Singh and Ravikumar, Pradeep Kumar and Bisk, Yonatan and Salakhutdinov, Ruslan},
  booktitle={International Conference on Learning Representations (ICLR)},
  year={2021}
}

@inproceedings{radford2021learning,
  title={Learning transferable visual models from natural language supervision},
  author={Radford, Alec and Kim, Jong Wook and Hallacy, Chris and Ramesh, Aditya and Goh, Gabriel and Agarwal, Sandhini and Sastry, Girish and Askell, Amanda and Mishkin, Pamela and Clark, Jack and others},
  booktitle={International Conference on Machine Learning (ICML)},
  pages={8748--8763},
  year={2021},
  organization={PMLR}
}

@inproceedings{kim2021vilt,
  title={ViLT: Vision-and-language transformer without convolution or region supervision},
  author={Kim, Wonjae and Son, Bokyung and Kim, Ildoo},
  booktitle={International Conference on Machine Learning (ICML)},
  pages={5583--5594},
  year={2021},
  organization={PMLR}
}

@inproceedings{zhou2022detecting,
  title={Detecting twenty-thousand classes using image-level supervision},
  author={Zhou, Xingyi and Girdhar, Rohit and Joulin, Armand and Kr{\"a}henb{\"u}hl, Philipp and Misra, Ishan},
  booktitle={European Conference on Computer Vision (ECCV)},
  pages={350--368},
  year={2022},
  organization={Springer}
}

@inproceedings{xu2022groupvit,
  title={GroupViT: Semantic segmentation emerges from text supervision},
  author={Xu, Jiarui and De Mello, Shalini and Liu, Sifei and Byeon, Wonmin and Breuel, Thomas and Kautz, Jan and Wang, Xiaolong},
  booktitle={IEEE/CVF Conference on Computer Vision and Pattern Recognition (CVPR)},
  pages={18134--18144},
  year={2022}
}

@inproceedings{gu2021open,
  title={Open-vocabulary object detection via vision and language knowledge distillation},
  author={Gu, Xiuye and Lin, Tsung-Yi and Kuo, Weicheng and Cui, Yin},
  booktitle={International Conference on Learning Representations (ICLR)}
}

@inproceedings{luddecke2022image,
  title={Image segmentation using text and image prompts},
  author={L{\"u}ddecke, Timo and Ecker, Alexander},
  booktitle={IEEE/CVF Conference on Computer Vision and Pattern Recognition (CVPR)},
  pages={7086--7096},
  year={2022}
}

@article{wang2021actionclip,
  title={ActionCLIP: Adapting language-image pretrained models for video action recognition},
  author={Wang, Mengmeng and Xing, Jiazheng and Mei, Jianbiao and Liu, Yong and Jiang, Yunliang},
  journal={IEEE Transactions on Neural Networks and Learning Systems},
  year={2023},
  publisher={IEEE}
}

@article{luo2022clip4clip,
  title={CLIP4Clip: An empirical study of CLIP for end to end video clip retrieval and captioning},
  author={Luo, Huaishao and Ji, Lei and Zhong, Ming and Chen, Yang and Lei, Wen and Duan, Nan and Li, Tianrui},
  journal={Neurocomputing},
  volume={508},
  pages={293--304},
  year={2022},
  publisher={Elsevier}
}

@inproceedings{brown2020language,
  title={Language models are few-shot learners},
  author={Brown, Tom and Mann, Benjamin and Ryder, Nick and Subbiah, Melanie and Kaplan, Jared D and Dhariwal, Prafulla and Neelakantan, Arvind and Shyam, Pranav and Sastry, Girish and Askell, Amanda and others},
  booktitle={Advances in Neural Information Processing Systems (NeurIPS)},
  volume={33},
  pages={1877--1901},
  year={2020}
}

@article{ramesh2022hierarchical,
  title={Hierarchical text-conditional image generation with CLIP latents},
  author={Ramesh, Aditya and Dhariwal, Prafulla and Nichol, Alex and Chu, Casey and Chen, Mark},
  journal={arXiv:2204.06125},
  year={2022}
}

@inproceedings{shafiullahclip,
  title={CLIP-Fields: Weakly supervised semantic fields for robotic memory},
  author={Shafiullah, Nur Muhammad Mahi and Paxton, Chris and Pinto, Lerrel and Chintala, Soumith and Szlam, Arthur},
  booktitle={ICRA2023 Workshop on Pretraining for Robotics (PT4R)},
  year={2023}
}
\end{document}